\documentclass[journal,compsoc,onecolumn]{IEEEtran}
\usepackage{amssymb}
\usepackage{amsmath,amsfonts}
\usepackage{amsthm}
\usepackage{algorithmic}
\usepackage{algorithm}
\usepackage{array}
\usepackage{textcomp}
\usepackage{epsfig}
\usepackage{stfloats}
\usepackage{url}
\usepackage{verbatim}
\usepackage{caption}
\usepackage{subcaption}
\usepackage{cite}
\usepackage{booktabs}
\usepackage[dvipsnames]{xcolor}
\usepackage{hyperref}
\usepackage{graphicx}
\usepackage[algo2e]{algorithm2e}
\usepackage{setspace}
\usepackage{etoolbox}
\usepackage{graphicx}
\usepackage{threeparttable}
\usepackage{mathrsfs}
\usepackage{multirow}
\usepackage{stmaryrd}
\usepackage{color,array}
\usepackage{graphicx}

\newtheorem{definition}{Definition}[section]
\newtheorem{assumption}{Assumption}
\newtheorem{theorem}{Theorem}

\newtheorem{lemma}{Lemma}

\newtheorem{remark}{Remark}

\begin{document}

\title{Topology-Driven Clustering: Enhancing Performance with Betti Number Filtration} 

\author{Arghya Pratihar, Kushal Bose, and Swagatam Das

\thanks{
Electronics and Communication Sciences Unit, Indian Statistical Institute, Kolkata, India \\ \\
E-mail: arghyapratihar24@gmail.com, kushalbose92@gmail.com, swagatam.das@isical.ac.in}
}

\maketitle

\begin{abstract}
Clustering aims at partitioning data points into groups of similar objects without knowing about the class labels. However, clustering datasets with complex geometric structures, such as nonconvex shapes, multiple scales, or intertwined manifolds, remains challenging for traditional algorithms that primarily rely on Euclidean or kernel-based similarity measures. Topological Data Analysis (TDA), particularly persistent homology, provides a powerful framework for capturing intrinsic structural properties of data, including connected components, loops, and higher-dimensional features across multiple scales. In this work, we propose a novel topological clustering algorithm called \textbf{Betti Number Filtration-based Topological Clustering (BFTC)}. The proposed method constructs local Vietoris–Rips filtrations around each data point and computes Betti numbers up to a prescribed dimension. These Betti numbers across filtration scales form \emph{Betti sequences}, which serve as multiscale topological signatures of local neighborhoods. By comparing Betti sequences among neighboring points, BFTC identifies topologically similar neighbors and refines the neighborhood graph to construct a topology-aware similarity structure and spectral clustering is applied to obtain the final clusters. Experimental results on several synthetic and real-world datasets demonstrate that BFTC effectively clusters complex and intertwined structures and consistently outperforms several state-of-the-art topology-based clustering methods.
\end{abstract}

\begin{IEEEkeywords}
Clustering, Topological Data Analysis, Vietoris-Rips Filtration, Betti Numbers, Persistent Homology, Outlier Removal.
\end{IEEEkeywords}

\section{Introduction}
\IEEEPARstart{T}opological clustering heralded a new learning paradigm that utilizes the concepts of topology to cluster data points. The harnessing of complex shapes and intricate structures embedded within the data is the key foundation of Topological Data Analysis (TDA) \cite{zomorodian2012tda}, \cite{tda}. In particular, persistent homology \cite{bois2024persistence} provides a robust framework for identifying connected components, loops, voids, and higher-dimensional features across multiple scales. Traditional clustering algorithms such as $k$-Means \cite{kmeans}, Spectral \cite{spectral}, Mean-Shift \cite{meanshift}, DBSCAN \cite{dbscan}, OPTICS \cite{optics}, and Agglomerative \cite{agglomerative} primarily rely on similarity measures derived from Euclidean or kernel-based distances. While effective in many applications, these approaches often fail to capture intrinsic topological structures, especially in datasets exhibiting nonconvex shapes, multiple scales, or higher-dimensional features such as loops and voids. To bridge this gap, several topologically informed clustering methods have been proposed. ToMATo \cite{tomato} integrates density-based mode seeking with persistence-guided merging, enabling cluster identification through $0$-dimensional homology of the density landscape. However, its reliance on connected components limits its ability to incorporate higher-dimensional topological information. TPCC \cite{tpcc}, on the other hand, constructs simplicial complexes and employs the $k^{\text{th}}$ Hodge Laplacian to cluster $k$-simplices, thereby incorporating higher-order connectivity. Nevertheless, TPCC may struggle to distinguish regions of a manifold where topological features emerge at different filtration scales. Recently, Topological $k$-Means (TKM) \cite{topokmeans} extends the classical $k$-Means framework by embedding persistence diagrams into reproducing kernel Hilbert spaces (RKHS). In this setting, cluster centroids are defined as Fr\'echet means of persistence-based representations, and clustering is achieved by minimizing topological variance within clusters. While TKM successfully incorporates global topological summaries into a centroid-based optimization framework, it primarily captures global persistence characteristics rather than localized multiscale topology.

Despite these advances, existing methods either focus on $0$-dimensional connectivity, rely on global persistence summaries, or operate at fixed topological scales. To address these limitations, we propose an efficient clustering algorithm named \textbf{B}etti Number \textbf{F}iltration-based \textbf{T}opological \textbf{C}lustering (\textbf{BFTC}). Our approach constructs Vietoris-Rips filtrations over local neighborhoods of each data point and computes persistent homology to extract Betti numbers up to a prescribed feature dimension. By comparing Betti sequences across filtration scales, we quantify topological similarity between neighboring points. This strategy enables BFTC to simultaneously capture local geometric proximity and higher-dimensional topological structures in a multiscale manner, leading to improved clustering performance on complex datasets.

Our method initially constructs a neighborhood graph using the $k$-nearest-neighbor \cite{knn} or $\varepsilon$-neighbor from the entire point set. Concurrently, we define Vietoris-Rips (VR) complexes for each point containing the set for a fixed filtration length. VR complexes, a well-adopted simplicial complex, are our primary choice due to their advantages in computational speed. Subsequently, for each point, we estimate the Betti number with respect to each VR complex for a fixed dimension. The maximum dimension can flexibly be chosen, which is the number of features of the data points. For a fixed dimension, we can arrange the Betti numbers for all VR complexes, which we termed as \textit{Betti sequence}. The length of the Betti sequence will be the length of the filtration. BFTC also offers a provision to select the maximum dimension to which we are interested in estimating Betti numbers. This provides the scope enabling flexibility for the choice of the dimension, which resembles the feature selection strategy prevailing in the machine learning paradigm. 

\begin{figure*}
    \centering
    \includegraphics[width=\textwidth]{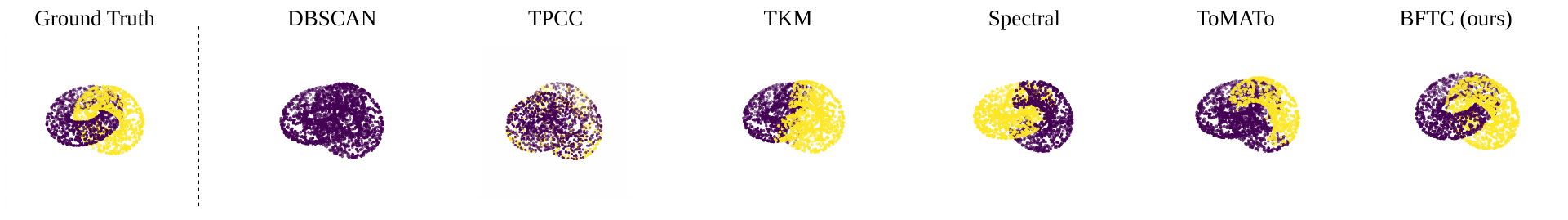}
    \caption{A comparative study of the different clustering algorithms DBSCAN, TPCC, TKM, Spectral, and ToMATo, with the proposed BFTC ($k$-NN\_BFTC), applied on the Linked-Tori dataset. The performance of BFTC is commendable compared with the rest of the clustering methods. }
    \label{fig:motivate_img}
\end{figure*}

\noindent
\textbf{Contributions.} 
Our contributions are outlined as follows:
\begin{itemize}
    \item Exploring topological characteristics of input data points, we proposed an efficient algorithm, BFTC, to perform data clustering. We identified Vietoris-Rips complexes of the points for a set of predetermined thresholds and evaluated Betti numbers up to a certain dimension for each complex. Later, Betti sequences are constructed by sequentially arranging Betti numbers for each VR-complex for every point, and the similarity of Betti sequences is considered to measure the topological similarity of the points. 
    \item Our proposed algorithm, BFTC, harnesses the underlying complex topological characteristics. Thus, BFTC is particularly effective for datasets containing complex shapes and structures intertwined with one another.  
    \item We offer an extensive theoretical underpinnings on the convergence analysis of BFTC. All the proofs and derivations are deferred to the Supplementary.
\end{itemize}

\section{Related Works}

Topological Data Analysis (TDA) \cite{wasserman2018topological} provides a versatile framework for extracting robust geometric and topological information from complex and high-dimensional data. A foundational tool in TDA is persistent homology, originally introduced in \cite{zomorodian2004computing, edelsbrunner2002topological}, which enables the multiscale characterization of topological features such as connected components, cycles, and higher-dimensional voids across varying resolution parameters.

Motivated by these capabilities, several clustering methods have been developed that explicitly incorporate topological information. An overview of how topological concepts can be integrated into clustering algorithms is provided in \cite{panagopoulos2022topological}. Among the most influential approaches is ToMATo (Topological Mode Analysis Tool) \cite{tomato}, which combines density-based mode seeking with a persistence-guided cluster-merging strategy. By analyzing the persistence of modes in the estimated density landscape, ToMATo is able to identify clusters while filtering out topological noise. Other approaches involve simplicial complexes and higher-order connectivity. TPCC \cite{tpcc} constructs a simplicial complex from point cloud data and performs clustering on $k$-simplices using the $k^{\text{th}}$ Hodge Laplacian, thereby capturing interactions beyond pairwise relationships. 
Topological $k$-Means (TKM) \cite{topokmeans} extends the classical $k$-Means clustering paradigm by incorporating persistent homology through reproducing kernel Hilbert space (RKHS) embeddings of persistence diagrams. While TKM effectively integrates global topological information into a centroid-based clustering framework, it requires the number of clusters to be specified a priori and primarily relies on global persistence representations rather than localized neighborhood topology. Several TDA-based clustering methods have also been proposed for structured data types. For instance, \cite{islambekov2019unsupervised} and \cite{majumdar2020clustering} develop persistent-homology-based clustering techniques for space--time and time-series data, respectively, while \cite{yuvarajtopological} introduces a TDA-driven framework for clustering complex multilayer networks. In contrast to the above approaches, our proposed method is built upon constructing Vietoris--Rips filtrations over local neighborhoods around each data point. Instead of relying on global persistence summaries or density landscapes, we compute persistent homology on these localized filtrations and extract Betti numbers up to a prescribed feature dimension. Topological similarity between a point and its neighbors is quantified by comparing the resulting Betti sequences across filtration scales. This similarity measure captures both geometric proximity and multiscale topological structure in local neighborhoods. Consequently, our algorithm, BFTC, jointly leverages local geometry and higher-dimensional topology, distinguishing it from global persistence-based methods such as TKM and connectivity-focused approaches such as NNVR \cite{bois2024persistence}.

\section{Proposed Method}

\subsection{Notations and Preliminaries.}
Assume a set of points lies in some metric space $(X,d)$, where $X$ denotes the point cloud and $d$ denotes the distance measure equipped in the metric space. Before delving into further details, let us define some domain-specific terminologies as follows.

\begin{definition}[Simplicial Complex] Let $X$ be a metric space. An abstract simplicial complex is a collection $K$ of finite subsets of $X$ such that $\sigma \in K$ and if for all $\tau \subseteq \sigma$ then $\tau \in K$. If $|\sigma|=k+1$ then $\sigma$ is called a $k$-simplex.\\
A 0-simplex is a point, a 1-simplex is an edge, a 2-simplex is a triangle, and a 3-simplex is a tetrahedron.
\end{definition}

A plethora of simplicial complexes prevails, such as Alpha \cite{alpha}, \v{C}ech \cite{cech}, Vietoris-Rips \cite{vrcomplex} complexes etc. The appropriate selection of the simplicial complexes hinges on several factors, viz., complexity of the underlying target tasks, characteristics of the input datasets, and cost for computing. Some simplicial complexes exhibit rich theoretical properties but are computationally intensive, while others are more efficient but less theoretically robust. However, Vietoris-Rips (VR) complexes provide a practical balance, emerging as an appropriate candidate for various applications and faster computational complexity. Therefore, we will resort to Vietoris-Rips complexes to construct our algorithm. 

\begin{definition}[Vietoris-Rips Complex] 
A Vietoris-Rips complex denoted by $VR(X,\varepsilon)$, is an abstract simplicial complex whose $k$-simplices consist of points whose pairwise distance is less than or equal to $\epsilon$. 
Let $(X, d)$ be a metric space and $\epsilon$ be a real parameter. Then $VR(X,\varepsilon)$ for all simplices $\sigma$ by :\\
$$
\begin{array}{r}
\forall \sigma \in K, \sigma \in V R(X, \varepsilon) \Leftrightarrow \forall x, y \in \sigma, d(x, y) \leq \varepsilon
\end{array}
$$
\end{definition}

\begin{definition}[Chain Complex]
    The vector space $C_k(K)$ is the formal sums of the k-simplices of $K$ with coefficients in $\mathbb{F}$, that is, if the k-simplices of $K$ are $\sigma_1, \sigma_2, \ldots \sigma_{n_k}$ then:
    $$
        c=\sum_{i=1}^{n_k} c_i \sigma_i ,\,\, c_i \in \mathbb{F}.
    $$
\end{definition}

\noindent
A chain complex is defined as,
$$
\cdots C_{k+1} \stackrel{\partial_{k+1}}{\longrightarrow} C_{k} \stackrel{\partial_{k}}{\longrightarrow} \cdots \stackrel{\partial_{2}}{\longrightarrow} C_{1} \stackrel{\partial_{1}}{\longrightarrow} C_{0} \stackrel{\partial_{0}}{\longrightarrow} 0.
$$

The images of $\partial_{k}$ are defined as \textit{boundaries} and the kernels of $\partial_{k-1}$ are defined as \textit{cycles}; $\operatorname{ker}\left(\partial_{k}\right)$ and $\operatorname{im}\left(\partial_{k+1}\right)$ are both subspaces of $C_{k}$ and we have, 
$$
B_{k}:=\operatorname{im}\left(\partial_{k+1}\right) \subseteq \operatorname{ker}\left(\partial_{k}\right)=: Z_{k}.
$$
The $k^{th}$ Homology group $H_k$ of a simplicial complex is defined as :
$$
H_{k}:=Z_{k} / B_{k}=\operatorname{ker}\left(\partial_{k}\right) / \operatorname{im}\left(\partial_{k+1}\right), \quad k=0,1, \ldots, d.
$$

\begin{definition}[Persistent Homology]
Persistent homology captures changes in the homology of a sequence of topological spaces. Given a filtered complex $K$, the persistent homology tracks the evolution of homology groups $H_n(K_i)$ over different levels of the filtration.
\end{definition}

To analyze the topological properties of the point cloud, we equip it with a simplicial complex structure and construct a series of nested simplicial complexes through filtration. This involves systematically varying a threshold parameter to build a sequence of simplicial complexes. This approach underscores the importance of persistent homology, a widely used method in topological data analysis. By examining the evolution of these complexes across the filtration, we derive key topological summaries, including the number of connected components, tunnels (1-dimensional holes), voids or cavities (2-dimensional holes), and, if any higher-dimensional features, captured through Betti numbers.

\begin{definition}[Betti Numbers] The $m^{\text{th}}$ Betti number $\beta_{m}, m \in Z^{+}$ denotes the dimension of the $m^{th}$ homology group of a simplicial complex. Intuitively $\beta_{m}$ counts the number of $m$-dimensional holes in a simplicial complex $K$.
\end{definition}
$$
\beta_{m}(K)=\operatorname{dim}\left(H_{m}(K)\right)=\operatorname{nullity}\left(\partial_{m}\right)-\operatorname{rank}\left(\partial_{m+1}\right).
$$

For example, given a complex, $\beta_{0}$ denotes the number of connected components; $\beta_{1}$ denotes the number of one-dimensional holes; $\beta_{2}$ denotes the number of two-dimensional holes.

\begin{definition}[Bottleneck Distance] \label{def:bottleneck}
 The Bottleneck distance \( d_B\) between two persistence diagrams, \( D_1 \) and \( D_2 \), is the infimum of the cost of all matchings, where the cost is the maximum distance between matched points.
\end{definition}
\[ d_B(D_1, D_2) = \inf_{\gamma \in \Gamma(D_1, D_2)} \sup_{x \in D_1} \| x - \gamma(x) \|_{\infty}, \]
where \(\Gamma(D_1, D_2)\) denotes the set of all bijections \(\gamma\) from \(D_1 \cup \Delta\) to \(D_2 \cup \Delta\), including mappings to the diagonal \(\Delta\) to adjust diagrams of different sizes.

\subsection{Our Proposed Algorithm: BFTC}
In this section, we present our proposed clustering algorithm, Betti Number Filtration-based Topological Clustering (BFTC) in detail.

\begin{algorithm}[h]
        \caption{BFTC Algorithm ($k$-NN\_BFTC or $\varepsilon\_$BFTC)}
        \label{main_algo}
        \scriptsize
        \begin{algorithmic}[1]
            \STATE
            \textbf{Input:} Data (X), Maximum dimension = $M$, Filtration Length = $L$, $k$ = number of $k$-nearest neighbors, Neighborhood radius $=\varepsilon$, $p$ = number of clusters.\\
            \vspace{0.5em}
            \STATE
            \textbf{Output:} Clusters $C_1, C_2 \ldots C_p$.
            \let \oldnoalign \noalign
            \let \noalign \relax
            \let \noalign \oldnoalign
            \vspace{0.7em}
            \STATE We construct a $k$-NN graph or $\varepsilon$-neighborhood graph where 
            $N_k(x_i)$ (or $N_{\varepsilon}(x_i)$) as defined in Eq. \eqref{eq:nbd} is the neighborhood of $x_i$. 
            \STATE Define scale sequence $\epsilon_{1}$, $\epsilon_{2}$, \ldots, $\epsilon_{L}$ equally spaced values from 0 to D, D is defined as \eqref{eq:knn} for $k$-NN or \eqref{eq:eps} for $\varepsilon$-neighborhood. \\
            \STATE
            \For{{$i$ from $1$ to $n$}}{
            \For{{$l$ from $1$ to $L$}}{Let $VR_l(x_i)$ be the Vietoris Rips complex corresponding to the threshold $\epsilon_{l}$ such that $VR_1(x_i) \subseteq VR_2(x_i) \subseteq \ldots \subseteq VR_L(x_i).$}}
            \STATE \For{{$i$ from $1$ to $n$}}{
            \For{{$m$ from $0$ to $M$ }}{
            $\beta_{m}^{(i)} =(\beta_{1m}^{(i)},\beta_{2m}^{(i)}, \ldots \beta_{Lm}^{(i)} )$\\}
            }
           \STATE \For{{$i$ from $1$ to $n$}}{
            \For{{$j$ from $1$ to $ind[i]$}}{
            $(d_{ij})^m= \frac{\langle \beta_m^{(i)} , \beta_m^{(j)} \rangle}{{\left\|\beta_m^{(i)}\right\|}{\left\|\beta_m^{(j)}\right\|}} \,\, m=0,\cdots, M$, where $ind[i]$ denotes the indices of the points in $\mathcal{N}(x_i)$.
            }
           }
           \STATE \For{{$i$ from $1$ to $n$}}{
           \For{{$j$ from $1$ to $ind[i]$}}{
           Check whether  $(d_{ij})^m \geq \alpha_{m} \,\, m=0,\cdots, M.$}
           }
           \STATE Retain $x_j$ in $\mathcal{N}(x_i)$ if condition holds; otherwise remove.
           \STATE Construct symmetric weighted adjacency matrix $\mathcal{A'}$: \\
            \For{{$i$ from $1$ to $n$}}{
            \For{{$j$ from $1$ to $n$}}{
            $$
            \mathcal{A}'_{ij} = exp\left({-\frac{\left\|x_i - x_j \right\|^2}{2{\sigma^2}}}\right) * a'_{ij} * a'_{ji}
            $$ as described in Eq. \eqref{eq:kernel}.
            }}
            \STATE Form the Laplacian Matrix $L'= D' -\mathcal{A'}$. Calculate the $p$-eigenvectors corresponding to the smallest $p$-eigenvalues of $L'$. 
            \STATE Apply $k$-Means clustering to obtain the clusters.
            \end{algorithmic}
    
\end{algorithm}

\vspace{3pt}
\noindent 
\textbf{Neighborhood Graph Construction.}
The first step of BFTC involves constructing a local neighborhood graph that captures geometric proximity among data points. Let $X = \{x_1, \dots, x_n\} \subset \mathbb{R}^d$ be the input dataset. We consider two alternative strategies for the construction of the neighborhood.
\begin{itemize}
    \item $k$-Nearest-Neighbor ($k$-NN) Graph.
    For a fixed integer $k$, define the neighborhood of $x_i$ as,
    \[
    N_k(x_i) = \{ \text{the } k \text{ nearest points to } x_i \text{ under } \|\cdot\| \}.
    \]
    An edge is established between $x_i$ and each $x_j \in N_k(x_i)$. After symmetrization, this forms the $k$-NN  graph. The $k$-NN construction guarantees that each node has at least $k$ local connections.
    \item $\varepsilon$-Neighborhood Graph. For a fixed radius $\varepsilon > 0$, define the neighborhood of $x_i$ as,
    \[
    N_{\varepsilon}(x_i) = \{ x_j \in X \mid \|x_i - x_j\| \le \varepsilon, \ j \neq i \}.
    \]

\end{itemize}

An edge is formed between $x_i$ and $x_j$ whenever the above condition holds. The resulting $\varepsilon$-graph connects points lying within a prescribed geometric scale and naturally aligns with metric-threshold-based constructions such as Vietoris--Rips complexes.
More generally, let $\mathcal{N}(x_i)$ denote a neighborhood operator defined by,
\begin{equation} \label{eq:nbd}
\mathcal{N}(x_i) =
\begin{cases}
N_k(x_i), & \text{for } k\text{-NN construction}, \\
N_{\varepsilon}(x_i), & \text{for } \varepsilon\text{-neighborhood construction}.
\end{cases}
\end{equation}
The initial adjacency matrix $A$ is then defined as,
\begin{equation} \label{eq:adj}
a_{ij} =
\begin{cases}
1, & x_j \in \mathcal{N}(x_i), \\
0, & \text{otherwise}.
\end{cases}
\end{equation}
This neighborhood graph serves as the geometric scaffold for constructing local Vietoris--Rips filtrations and subsequent Betti sequence computation. Neighborhood construction provides the local geometric structure upon which persistent homology is computed. The $k$-NN graph enforces uniform cardinality and controlled sparsity, offering robustness under heterogeneous density and bounded computational complexity. In contrast, the $\varepsilon$-neighborhood graph enforces a fixed geometric scale, aligning naturally with Vietoris--Rips filtrations and scale-based topology extraction. By adopting a unified formulation, BFTC accommodates both constructions, allowing flexibility based on dataset characteristics and computational considerations. We will consider two variants: $k$-NN\_BFTC for $k$-Nearest Neighbor and $\varepsilon\_$BFTC for $\varepsilon$ neighborhood graph based on the choice of neighborhood.

\vspace{3pt}
\noindent
\textbf{Estimating Betti Numbers for VR-filtration.} At this stage, we will incorporate VR-filtration and compute Betti numbers of the entire point set. The filtration scale depends on the chosen neighborhood strategy:

\begin{itemize}
    \item \textbf{$k$-NN case:}  
    Let $d_k(x_i)$ denote the distance from $x_i$ to its $k$-th nearest neighbor. Define
    \begin{equation} \label{eq:knn}
    D = \max_{i} d_k(x_i).
    \end{equation}
    
    \item \textbf{$\varepsilon$-neighborhood case:}  
    The neighborhood radius $\varepsilon$ naturally determines the maximum filtration scale, hence
    \begin{equation} \label{eq:eps}
    D = \varepsilon.
    \end{equation}
\end{itemize}
Considering the filtration length $L$, we divide $L$ equally spaced points $\epsilon_{1}$, $\epsilon_{2}$, \ldots $\epsilon_{L}$ in the range of $0$ to $D$.
For each point $x_i \in X$ and for each scale $\varepsilon_\ell$, we construct a local Vietoris--Rips complex:
$
VR_\ell(x_i) = VR\big(\mathcal{N}(x_i) \cup \{x_i\}, \varepsilon_\ell\big),
$
such that
\[
VR_1(x_i) \subseteq VR_2(x_i) \subseteq \dots \subseteq VR_L(x_i).
\]
This construction captures multiscale topological features of the local geometric neighborhood of $x_i$. If the number of features of the data points is $d$, then we will compute Betti numbers for each $VR_{l}(x_i)$ for each dimension ranging from $0$ to $M$ where $M \in [0, d-1], \forall x_i \in X$. Therefore, $\beta^{(i)}_{lm}$ denotes the $m^{th}$ Betti number of the corresponding VR-complex $VR_{l}(x_i)$ for the point $x_i$, where $m \in [0, M]$. \\
\noindent
\begin{remark}
    Although homology groups are formally defined up to the ambient feature dimension of the data, the maximum homology dimension $M$ is treated as a user-specified parameter independent of the feature dimension. Due to the combinatorial growth of Vietoris--Rips complexes, persistent homology is typically computed only in low dimensions. In our experiments:

\begin{itemize}
    \item For synthetic 3D geometric datasets, we restrict to $M \le 2$. as higher dimensional homology groups vanish.
    \item For selected real-world datasets (e.g., Zoo, Ecoli, Glass), we compute up to $M \le 5$ when computationally feasible.
\end{itemize}
Importantly, $M$ does not scale with the ambient feature dimension; rather, it reflects the intrinsic topological complexity that we are aiming to capture.
\end{remark}

\vspace{3pt}
\noindent
\textbf{Computing Similarity between Betti sequences.}
Let us consider Betti numbers up to dimension $M$ where $M \in [0, d-1]$, then for each dimension $m \in [0, M]$ compute the Betti numbers of $VR_{l}(x_i), \,\, \forall l \in [1, L]$. Therefore, we have $L$ different Betti numbers for a fixed dimension $m$ for each point $x_i$. Those Betti numbers formed a sequence which is represented as $\beta^{(i)}_{m} = \{\beta^{(i)}_{1m}, \beta^{(i)}_{2m}, \cdots, \beta^{(i)}_{Lm}\} \in \mathbb{R}^{L}$ for $L$ different $VR$-complexes. This sequence is termed as \textit{Betti sequence}. We now have $(M+1)$ Betti sequences for each point in $X$. The Betti sequences will be employed to measure the similarity scores between pairs of adjacent points in the neighborhood graph as follows,
\begin{equation} \label{eq:cosine}
    (d_{ij})^{m}= \frac{\langle \beta_m^{(i)}, \beta_m^{(j)} \rangle}{{\left\|\beta_m^{(i)}\right\|}{\left\|\beta_m^{(j)}\right\|}}, \,\, m=0,\cdots, M.
\end{equation}
where $\beta_{m}^{(i)}, \beta_{m}^{(j)}$ are the Betti sequence for $m$-th dimension of the $i^{th}$ and $j^{th}$ points respectively. The cosine similarity in Eq. \ref{eq:cosine} between the Betti sequences ensures the incorporation of topologically aware information into the neighborhood of the existing neighborhood graph from the data points.

\begin{remark}
BFTC extends the integration of persistent homology to clustering by incorporating Betti numbers up to the data's intrinsic dimension, rather than restricting to $\beta_0$ and $\beta_1$ as in CBN \cite{islambekov2019unsupervised}. These Betti numbers across filtration scales, which we term \emph{Betti sequences}, serve as multiscale topological signatures that capture richer structural information across filtration scales. By incorporating higher-dimensional Betti numbers and exploiting their multiscale behavior, BFTC provides a richer and more expressive representation of the underlying topology of the data. Furthermore, BFTC adopts a more flexible neighborhood construction strategy by supporting
both $k$-nearest neighbor and $\varepsilon$-neighborhood graphs, allowing the algorithm to adapt to datasets with sparsity and geometric structures. BFTC constructs a \emph{topology-aware weighted graph} and employs spectral clustering,
thereby integrating local topological similarity with the global graph structure. These enhancements enable BFTC to capture richer topological information and improve cluster separability in datasets containing complex or intertwined structures. 

\end{remark}

\begin{table*}[!ht]
\centering
\caption{The performances of two variants of BFTC on well-adopted six synthetic datasets, where the first three of them are $2$D, whereas the last three are synthetic $3$D. Additionally, we present results for Linked Tori with a noise level of $0.3$. BFTC outperformed all contenders on each dataset in terms of ARI and NMI, even with the presence of noise ($\rho$). The best and second-best results are highlighted in boldface and underlined, respectively.} 
\label{tab:synth_1}
\renewcommand{\arraystretch}{1.4}
\resizebox{\columnwidth}{!}{
\begin{tabular}{lcccccccccccc|cc}
\toprule
Methods / Datasets & \multicolumn{2}{c}{Cure-t2-4k} & \multicolumn{2}{c}{Smile1} & \multicolumn{2}{c}{3 MC} & \multicolumn{2}{c}{2 Sphere 2 Circle} & \multicolumn{2}{c}{Linked Tori} & \multicolumn{2}{c}{Torus Sphere Line} & \multicolumn{2}{|c}{Linked Tori ($\rho=0.3$)}\\
\midrule
 & ARI & NMI & ARI & NMI & ARI & NMI & ARI & NMI & ARI & NMI & ARI & NMI & ARI & NMI \\
\cmidrule{2-15} 
$k$-Means \cite{kmeans} & 0.43 & 0.66 & 0.54 & 0.60 & 0.77 & 0.78 & 0.81 & 0.79 & 0.35 & 0.32 & 0.73 & 0.77 & 0.17 & 0.13\\
Spectral \cite{spectral} & 0.86 & 0.89 & 0.68 & 0.76 & 0.70 & 0.75 & 0.84 & 0.81 & {0.91} & \underline{0.94} & 0.82 & 0.87 & {0.34} & {0.41} \\
Agglomerative \cite{agglomerative} & 0.46 & 0.69 & 0.86 & 0.91 & \textbf{1.00} & \textbf{1.00} & 0.86 & 0.83 & 0.43 & 0.42 & 0.65 & 0.67 & 0.23 & 0.22\\
Mean-Shift \cite{meanshift} & 0.32 & 0.38 & 0.33 & 0.47 & 0.58 & 0.66 & 0.74 & 0.73 & 0.14 & 0.11 & 0.82 & 0.79 & 0.01 & 0.02\\
DBSCAN \cite{dbscan} & 0.78 & 0.83 & \textbf{1.00} & \textbf{1.00} & 0.51 & 0.70 & 0.30 & 0.47 & 0.02 & 0.03 & 0.28 & 0.37 & 0.01 & 0.02 \\
Optics \cite{optics} & 0.70 & 0.76 & 0.56 & 0.71 & 0.32 & 0.53 & 0.19 & 0.35 & 0.49 & 0.41 & 0.42 & 0.45 & 0.11 & 0.17 \\
\midrule
ToMATo \cite{tomato} & \underline{0.87} & \underline{0.89} & 0.77 & 0.85 & \underline{0.91} & \underline{0.96} & 0.77 & 0.80 & 0.82 & 0.86 & 0.66 & 0.69 & 0.17 & 0.20 \\
TPCC \cite{tpcc} & 0.71 & 0.76 & 0.85 & 0.91 & 0.87 & 0.91 & \textbf{0.97} & \underline{0.91} & 0.31 & 0.36 & 0.81 & 0.86 & 0.13 & 0.21 \\
CBN \cite{islambekov2019unsupervised} & 0.68 & 0.77 & 0.62 & 0.68 & 0.67 & 0.81 & 0.54 & 0.68 & 0.52 & 0.47 & 0.56 & 0.51 & 0.10 & 0.14 \\
TKM \cite{topokmeans} & 0.67 & 0.75 & 0.72 & 0.81 & {0.61} & {0.76} & 0.79 & 0.85 & 0.54 & 0.61 & 0.89 & 0.92 & 0.31 & 0.37 \\
\midrule
\textbf{$k$-NN\_BFTC (Ours)} & \textbf{0.91} & \textbf{0.93} & \textbf{1.00} & \textbf{1.00} & \textbf{1.00} & \textbf{1.00} & \underline{0.96} & \textbf{0.94} & \textbf{1.00} & \textbf{1.00} & \underline{0.95} & \underline{0.97} & \textbf{0.49} & \textbf{0.56}\\

\textbf{$\varepsilon\_$BFTC (Ours)} & {0.84} & {0.87} & \underline{0.94} & \underline{0.97} & {0.88} & {0.91} & {0.93} & 0.89 & \underline{0.96} & 0.93 & \textbf{0.98} & \textbf{0.99} & \underline{0.42} & \underline{0.50}\\
\bottomrule
\end{tabular}}
\end{table*}

\vspace{3pt}
\noindent
\textbf{Removing Outliers and Updating Neighborhoods.}
The estimated cosine similarity scores will be employed to determine topologically similar neighbors. For each $x_i \in X$, let us define a set of thresholds $\alpha_0, \alpha_1, \alpha_2, \cdots, \alpha_M$ acting as the lower bounds for the $(M+1)$ cosine similarity scores $(d_{ij})^0, (d_{ij})^1, (d_{ij})^{2}, \cdots, (d_{ij})^M$ respectively. The thresholds are thereby applied to update the neighborhoods of $x_i$ as follows,
\begin{equation} \label{eq:neighbor}
    \mathcal{N'}(x_i) = \begin{cases}
        \mathcal{N}(x_i) \cap \{x_j\}, & (d_{ij})^{m} \ge \alpha_m, \forall \, m \in[0,M] \\
        \mathcal{N}(x_i) \setminus \{x_j\}, & \text{otherwise}
    \end{cases}
\end{equation}
where $\mathcal{N}(x_i)$ is the initial neighborhood graph. In this way, the unnecessary neighbors are eliminated, and the new neighborhood graph, $\mathcal{N'}(x_i)$ as constructed in Eq.\eqref{eq:neighbor}, contains comparatively more topologically informative neighbors. Notably, the defined thresholds are shared across all data points in the dataset. This solution removes outliers and unnecessary neighbors, yielding a sparser, more topologically informative graph.   

\begin{figure}
    \centering
    \includegraphics[width= 0.6\textwidth]{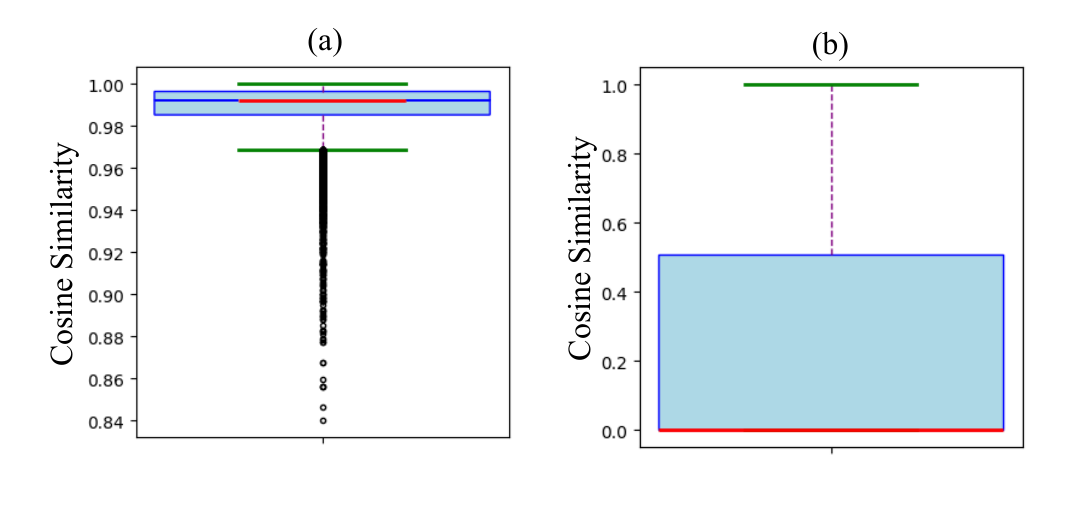}
    \caption{The box plots corresponding to cosine similarity scores for (a) $\beta_0$ and (b) $\beta_1$ are presented respectively for the Linked Tori dataset.}
    \label{fig:boxplot_vs_betti}
\end{figure}

\vspace{3pt}
\noindent
\textbf{Forming Topology-aware Graph and Finding Clusters.}
In the end, we define the adjacency matrix, $A'$ by further refining $A$ in Eq. \eqref{eq:adj} such that 
\[
a'_{ij} =
\begin{cases}
1, & x_j \in \mathcal{N'}(x_i), \\
0, & \text{otherwise}.
\end{cases}
\]
Note that $A'$ may not be necessarily symmetric; for that, we convert $A'$ into a symmetric weighted adjacency matrix $\mathcal{A'}$, assigning weights by imparting a Gaussian kernel in the following way,
\begin{equation} \label{eq:kernel}
    (\mathcal{A'})_{ij} = \begin{cases}
         \exp\left({-\frac{\left\|x_i - x_j \right\|^2}{2{\sigma^2}}}\right)* a'_{ij} * a'_{ji}, & \forall \, (a'_{ij} \land a'_{ji}) = 1 \\
         0, & \text{otherwise}
    \end{cases}
\end{equation}
where, $\sigma$ denotes the standard deviation of the data. The degree of each point $x_i \in X$ is denoted by ${d_i}' = \sum_{j=1}^{n} (\mathcal{A'})_{ij}$. The degree matrix $D'$ is the diagonal matrix. We construct the Laplacian matrix $L'=(D' -\mathcal{A'})$. Then we calculate the $p$ eigenvectors corresponding to the smallest $p$ eigenvalues by spectral decomposition of $L'$ and apply $k$-Means clustering to get the $p$ clusters. 

The pseudocode of BFTC with two variants is demonstrated in the Algorithm \ref{main_algo}.

\subsection{Complexity Analysis}
In this section, we are going to present the step-by-step complexity analysis of BFTC under the unified Neighborhood construction.

\vspace{3pt}
\noindent
\textbf{Neighborhood Graph Construction.} The first step involves constructing the neighborhood graph $\mathcal{N}(x_i)$.
\begin{itemize}
    \item $k$-NN Construction. Computing pairwise distances requires $\mathcal{O}(n^2 M)$ operations. Selecting the $k$-nearest neighbors for each point requires sorting or partial selection, leading to $\mathcal{O}(n \log n)$ for each point. Thus, the complexity is: $\mathcal{O}(n^2 M + n^2 \log n).$

    \item $\varepsilon$-Neighborhood Construction. Determining $\varepsilon$-neighbors requires computing all pairwise distances and checking whether $\|x_i - x_j\| \le \varepsilon$, which costs $\mathcal{O}(n^2 M).$
\end{itemize}

\vspace{3pt}
\noindent
\textbf{Estimating Betti Numbers for VR-Filtration.}
For each point $x_i$, persistent homology is computed on the local neighborhood $\mathcal{N}(x_i)\cup\{x_i\}$ across $L$ filtration scales up to homology
dimension $M$. In the $k$-NN variant, $D$ is determined by the maximum $k$-th nearest neighbor distance among all points, which requires $\mathcal{O}(n)$. In the $\varepsilon$-neighborhood variant, the radius $\varepsilon$ directly determines the maximum filtration scale, so in $\mathcal{O}(1)$.

For a filtration of length $L$, the Betti numbers of $L$ Vietoris--Rips
Complexes are computed for each point. Using efficient persistent homology implementations such as Ripser
\cite{ripser}, the computation for each point requires approximately $\mathcal{O}(|\mathcal{N}(x_i)|^{M}L)$ operations. For the $k$-NN variant, the total
computational cost becomes
$
\mathcal{O}(n k^{M} L).
$
For the $\varepsilon$-neighborhood variant, letting $\bar{N}$ denote the
average neighborhood size, the total complexity becomes
$
\mathcal{O}(n \bar{N}^{M} L).
$

\vspace{3pt}
\noindent
\textbf{Computing Similarity between Betti Sequences.}
Each Betti sequence $\beta_m^{(i)}$ has length $L$. Computing the cosine similarity between two Betti sequences therefore requires $\mathcal{O}(L)$ operations. Since the similarity is evaluated across all homology dimensions $m=0,\ldots, M$, the time complexity becomes $\mathcal{O}((M+1)L)$. For the $k$-NN variant, the total complexity is
$
\mathcal{O}(n k (M+1)L).
$
For the $\varepsilon$-neighborhood variant, the time complexity becomes
$
\mathcal{O}(n \bar{N} (M+1)L).
$

\begin{table*}[!ht]
\centering
\caption{The performances of two variants of BFTC on $7$ real-world benchmark datasets are presented. BFTC outperformed almost all contenders on all datasets in terms of ARI and NMI. The best and second-best results are highlighted in boldface and underlined, respectively.}
\renewcommand{\arraystretch}{1.4}
\resizebox{\columnwidth}{!}{\begin{Huge}
\scriptsize
\label{tab:real}
\begin{tabular}{lcccccccccccccc}
\toprule
Methods / Datasets & \multicolumn{2}{c}{Zoo} & \multicolumn{2}{c}{Ecoli}  &\multicolumn{2}{c}{Wisconsin B.C.} & \multicolumn{2}{c}{Phishing URL (5k)} & \multicolumn{2}{c}{Glass} & \multicolumn{2}{c}{ORHD} & \multicolumn{2}{c}{MNIST (5k)} \\
\midrule
 & ARI & NMI & ARI & NMI & ARI & NMI & ARI & NMI & ARI & NMI & ARI & NMI & ARI & NMI \\
\cmidrule{2-15} 
$k$-Means \cite{kmeans} & 0.43 & 0.65 & 0.41 & 0.55 & 0.47 & 0.45 & 0.01 & 0.02 & 0.21 & 0.35 & 0.55 & 0.69 & 0.29 & 0.44 \\
Spectral \cite{spectral} & 0.49 & \underline{0.67} & 0.38 & 0.54 & 0.41 & 0.42 & {0.18} & {0.24} & 0.15 & 0.32 & \underline{0.73} & \textbf{0.85} & 0.43 & {0.53} \\
Agglomerative \cite{agglomerative} & 0.44 & 0.63 & {0.49} & {0.57} & 0.29 & 0.32 & 0.10 & 0.16 & 0.26 & 0.39 & 0.69 & 0.77 & 0.37 & 0.46 \\
Mean-Shift \cite{meanshift} & 0.39 & 0.54 & 0.42 & 0.41 & 0.41 & 0.48 & 0.01 & 0.02 & \underline{0.27} & \underline{0.44} & 0.03 & 0.04 & 0.01 & 0.02 \\
DBSCAN \cite{dbscan} & 0.05 & 0.12 & 0.18 & 0.32 & 0.13 & 0.18 & 0.10 & 0.19 & 0.11 & 0.25 & 0.10 & 0.25 & 0.12 & 0.27 \\
Optics \cite{optics} & 0.34 & 0.64 & 0.25 & 0.28 & 0.10 & 0.18 & 0.05 & 0.16 & 0.13 & 0.29 & 0.01 & 0.02 & 0.01 & 0.01 \\
\midrule
ToMATo \cite{tomato} & 0.46 & 0.61 & 0.44 & 0.57 & 0.41 & 0.40 & 0.03 & 0.04 & 0.12 & 0.28 & 0.71 & 0.79 & 0.43 & 0.51 \\
TPCC \cite{tpcc} & 0.28 & 0.43 & 0.41 & 0.46 & 0.32 & 0.43 & 0.12 & 0.19 & 0.23 & 0.38 & 0.54 & 0.65 & 0.21 & 0.31 \\
CBN \cite{islambekov2019unsupervised} & 0.41 & 0.63 & 0.28 & 0.43 & 0.16 & 0.23 & 0.06 & 0.11 & 0.20 & 0.31 & 0.35 & 0.47 & 0.08 & 0.15 \\
TKM \cite{topokmeans} & {0.54} & 0.61 & 0.25 & 0.35 & \underline{0.52} & {0.58} & 0.05 & 0.08 & 0.18 & 0.25 & 0.64 & 0.72 & 0.14 & 0.29 \\
\midrule
\textbf{$k$-NN\_BFTC (Ours)} & \underline{0.62} & \underline{0.73} & \textbf{0.63} & \textbf{0.68} & \textbf{0.55} & \textbf{0.61} & \textbf{0.29} & \textbf{0.36} & \textbf{0.30} & \textbf{0.47}\ & \textbf{0.75} & \underline{0.81} & \textbf{0.47} & \underline{0.53} \\

\textbf{$\varepsilon\_$BFTC (Ours)} & \textbf{0.70} & \textbf{0.76} & \underline{0.61} & \underline{0.64} & {0.51} & \underline{0.59} & \underline{0.25} & \underline{0.32} & 0.25 & 0.39 & 0.72 & 0.78 & \underline{0.45} & \textbf{0.55} \\
\bottomrule
\end{tabular}
\end{Huge}}
\end{table*}

\vspace{3pt}
\noindent
\textbf{Removing Outliers and Updating Neighborhoods.}
Computing the updated neighborhood $\mathcal{N}'(x_i)$ requires comparing each edge's $(M+1)$ cosine similarity scores against the corresponding thresholds $\alpha_0, \ldots, \alpha_M$. Since there are at most $n \cdot |\mathcal{N}(x_i)|$ edges in the initial graph and the threshold comparison is $\mathcal{O}(M)$ per edge, the total cost is
$
\mathcal{O}(nkM),
$
for the $k$-NN variant and $\mathcal{O}(n\bar{N} M)$ for $\varepsilon$-neighborhood variant.

\vspace{3pt}
\noindent
\textbf{Forming the Topology-aware Graph and Finding Clusters.}
Updating the adjacency relations, therefore, requires \\
$\mathcal{O}\!\left(\sum_{i=1}^{n} |\mathcal{N}(x_i)|\right)$ operations.
For the $k$-NN variant, the refinement step
requires
$
\mathcal{O}(nk).
$
For the $\varepsilon$-neighborhood variant, the cost becomes
$
\mathcal{O}(n\bar{N}).
$
Next, the symmetric weighted adjacency matrix $\mathcal{A}'$ is constructed using the Gaussian kernel defined in Eq.~\eqref{eq:kernel}. Once the adjacency matrix is formed, the Laplacian matrix
$L' = D' - \mathcal{A}'$ is constructed and spectral clustering is performed.
Extracting the smallest $p$ eigenvalues and corresponding eigenvectors of $L'$
using the Lanczos algorithm \cite{lanczos} requires approximately $\mathcal{O}(pnk)$
operations for the $k$-NN graph and
$
\mathcal{O}(pn\bar{N})
$
for the $\varepsilon$-neighborhood graph, since the graph is sparse. Typically, $p \ll n$, the cost of this step is dominated by the number of edges of the graph.

\medskip
\noindent
\textbf{Overall Complexity.}
Combining all steps of the algorithm, the overall complexity becomes, for \textbf{$k$-NN\_BFTC:} $\mathcal{O}(n^2 M + n^2\log n + nk^{M}L + nk(M+1)L + nkM  + pnk)
$
and \textbf{$\varepsilon$\_BFTC:}
$
\mathcal{O}(n^2 M + n\bar{N}^{\,M}L + n\bar{N}(M+1)L + n\bar{N}M + pn\bar{N}).
$
Since $k, \bar{N}, L,$ and $M \ll n$, the dominant cost is the neighborhood construction stage,
leading to an overall complexity of approximately $\mathcal{O}(n^2 \log n)$ for the $k$-NN variant, and $\mathcal{O}(n^2)$ for the $\varepsilon$-neighborhood variant. 

\begin{figure}
    \centering
    \includegraphics[width= 0.45\textwidth, height= 0.35\textheight]{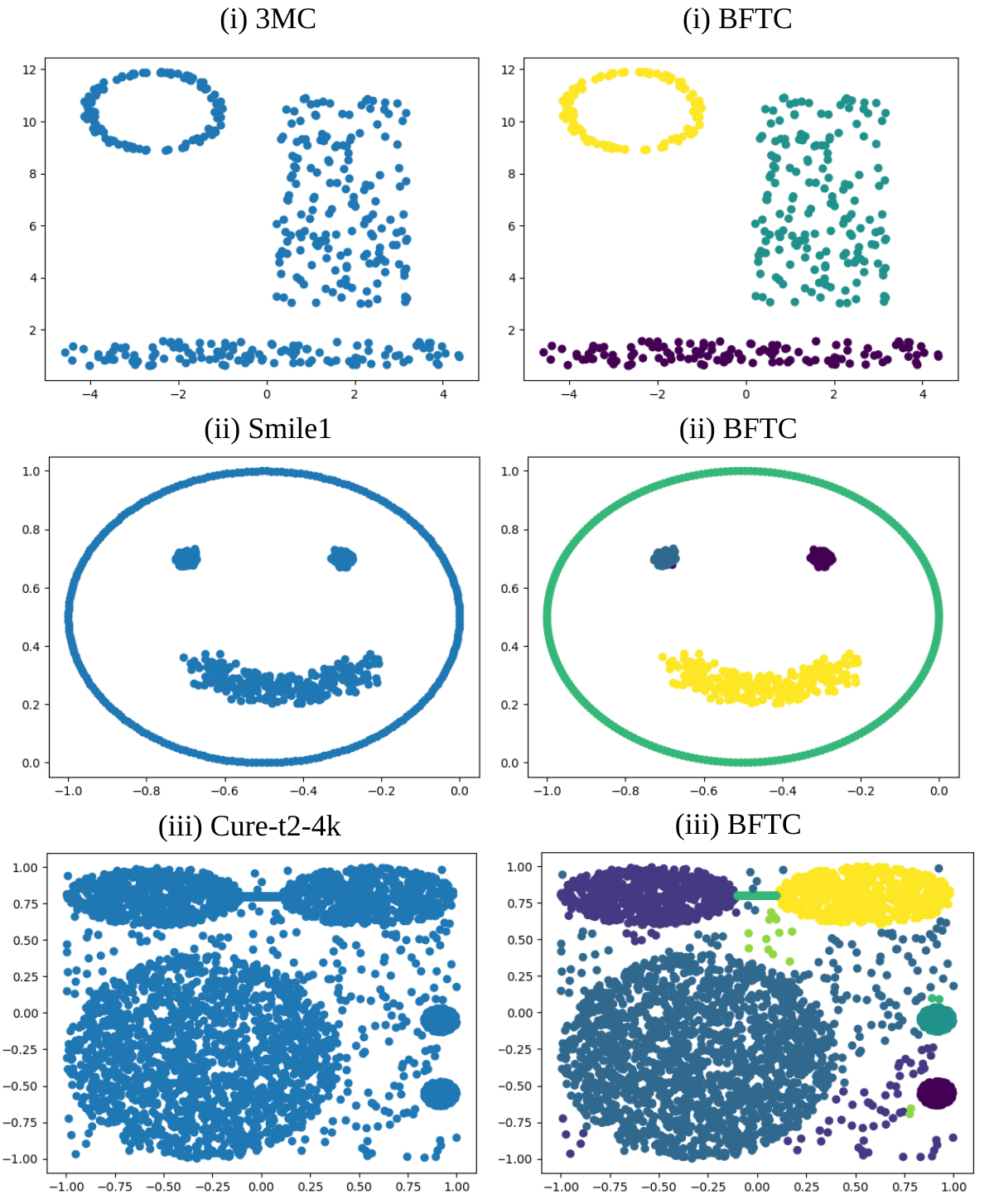}
    \caption{Performance of $k$-NN\_BFTC (ours) for $3$ Synthetic $2$D datasets 3MC, Smile1, Cure-t2-4k taken from clustering benchmark dataset.}
    \label{fig:synthetic_2d}
\end{figure}

\section{Theoretical Analysis} \label{sec:convergence}

We analyze the convergence properties of the proposed \emph{Betti Number Filtration-based Topological Clustering (BFTC)} algorithm.

\begin{assumption}[Finite Metric Sample]
\label{ass:finite}
Let $X_n = \{x_1,\ldots,x_n\} \subset (X,d)$ be a finite subset of a proper metric space.
\end{assumption}

\begin{assumption}[Generic Filtration Position]
\label{ass:generic}
No critical value (birth or death time) of the Vietoris--Rips filtration coincides with
any filtration scale $\{\varepsilon_\ell\}_{\ell=1}^L$.
\end{assumption}

\begin{assumption}[Threshold Separation]
\label{ass:threshold}
For each Betti dimension $m \in \{0,\ldots,M\}$, there exists $\eta_m > 0$ such that
\[
\bigl| (d_{ij})^m - \alpha_m \bigr| \ge \eta_m
\quad \text{for all neighbor pairs } (i,j).
\]
\end{assumption}

\begin{assumption}[Spectral Gap]
\label{ass:spectral}
Let $0 = \lambda_1 \le \lambda_2 \le \cdots \le \lambda_n$ be the eigenvalues of the
topology-aware graph Laplacian $L'$. There exists $\gamma > 0$ such that
\[
\lambda_{p+1} - \lambda_p \ge \gamma .
\]
\end{assumption}

\begin{lemma}[Stability of Vietoris--Rips Persistence]
\label{lem:vrstability}
Let $X_n, Y_n \subset X$ satisfy $d_H(X_n,Y_n) \le \delta$.
Then the corresponding persistence diagrams satisfy
\[
d_B(\mathrm{PH}(X_n), \mathrm{PH}(Y_n)) \le 2\delta ,
\]
where $d_B$ denotes the bottleneck distance as definition \ref{def:bottleneck}.
\end{lemma}

\begin{lemma}[Local Constancy of Betti Sequences]
\label{lem:betti}
Under Assumption~\ref{ass:generic}, there exists $\delta_0 > 0$ such that if
$d_H(X_n,Y_n) < \delta_0$, then
\[
\beta^{(i)}_m(X_n) = \beta^{(i)}_m(Y_n)
\quad \forall i,\; m .
\]
\end{lemma}

\begin{lemma}[Invariance of Refined Neighborhoods]
\label{lem:neighborhood}
Under Assumption~\ref{ass:threshold}, there exists $\delta_1 > 0$ such that if
$d_H(X_n,Y_n) < \delta_1$, then
\[
\mathcal{N}'(x_i;X_n) = \mathcal{N}'(x_i;Y_n) \quad \forall i .
\]
\end{lemma}

\begin{lemma}[Adjacency Matrix Stability]
\label{lem:adjacency}
Let $\mathcal{A}'(X_n)$ and $\mathcal{A}'(Y_n)$ be the Gaussian-weighted adjacency matrices defined by
Eq. \ref{eq:kernel}.
Then there exists a constant $C > 0$ such that
\[
\|\mathcal{A}'(X_n) - \mathcal{A}'(Y_n)\|_2 \le C\, d_H(X_n,Y_n)
\]
whenever $d_H(X_n,Y_n) < \delta, \quad \delta = \min(\delta_0,\delta_1)$.
\end{lemma}

\begin{lemma}[Spectral Subspace Convergence]
\label{lem:spectral}
Let $L_X := L'(X_n)$ and $L_Y := L'(Y_n)$ be the
topology-aware graph Laplacians corresponding to
point clouds $X_n$ and $Y_n$.
Assume:
\begin{enumerate}
    \item $\|L_X - L_Y\|_2 \le \varepsilon$,
    \item The eigenvalues of $L_X$ satisfy a spectral gap as in Assumption \ref{ass:spectral}.
    \[
    \lambda_{p+1}(L_X) - \lambda_p(L_X) \ge \gamma > 0.
    \]
\end{enumerate}

Let $U_X$ and $U_Y$ denote the matrices whose columns span
the eigenspaces corresponding to the first $p$ eigenvalues of
$L_X$ and $L_Y$, respectively.
Then
\[
\|\sin \Theta(U_X, U_Y)\|_2
\le
\frac{\varepsilon}{\gamma}.
\]
In particular, if
\[
\|L_X - L_Y\|_2 \le C\, d_H(X_n,Y_n),
\]
then
\[
\|\sin \Theta(U_X, U_Y)\|_2
\le
\frac{C}{\gamma}
\, d_H(X_n,Y_n).
\]
\end{lemma}

\begin{lemma} \label{lem:kmeans}
The $k$-Means algorithm applied to the spectral embedding converges in finitely many
iterations to a local minimum.
\end{lemma}

\begin{theorem}
\label{thm:convergence}
Under Assumptions~\ref{ass:finite}--\ref{ass:spectral}, there exists $\delta^\star > 0$
such that if $d_H(X_n,Y_n) < \delta^\star$, then BFTC produces identical clusterings on
$X_n$ and $Y_n$ up to permutation of labels. Moreover, the algorithm terminates in finite
time and converges to a local minimum of its final $k$-Means objective.
\end{theorem}

\begin{remark}
The preceding lemmas and Theorem~\ref{thm:convergence} establish the \emph{stability and convergence} of BFTC under small perturbations of the input data. In particular, small changes in the point cloud lead to stable persistence diagrams, locally invariant Betti sequences, unchanged refined neighborhoods, and stable
topology-aware adjacency matrices. Consequently, the corresponding spectral embedding also varies continuously whenever a spectral gap exists, while the final $k$-Means step converges in finite time. These results imply that BFTC is not only computationally finite, but also \emph{structurally stable}: small perturbations of the input dataset do not alter the resulting clustering, up to label permutation. Theorem~\ref{thm:convergence} therefore establishes strong convergence in the sense that, under reasonable geometric and spectral separation conditions, the entire topology-aware clustering pipeline is locally invariant. This provides a theoretical justification for the empirical robustness of BFTC observed in noisy and perturbed datasets.
\end{remark}

\section{Experiments}

\subsection{Details of the Datasets}
We validate the efficacy of two variants of BFTC on synthetic $2$D and $3$D datasets and some real-world datasets. Phishing URL, Zoo, Wisconsin Breast Cancer (Diagnostic), ORHD (Optical Recognition of Handwritten Digits), Glass, Ecoli, and MNIST datasets are taken from the UCI machine learning repository \cite{dua2017uci}. Cure-t2-4k, 3MC, and Smile1 are taken from Clustering Benchmark datasets available at https://github.com/deric/clustering-benchmark. For synthetic $3$D datasets, we have generated the Linked Tori dataset using $4000$ points ($2000$ points for each torus), Torus Sphere Line using $3300$ points ($2000$ for torus, $1000$ for sphere, $300$ for line), 2 Sphere 2 Circle using $3000$ points ($1000$ for each sphere and $500$ for each circle). 

\begin{figure}
    \centering
    \includegraphics[width=0.45\textwidth, height= 0.37\textheight]{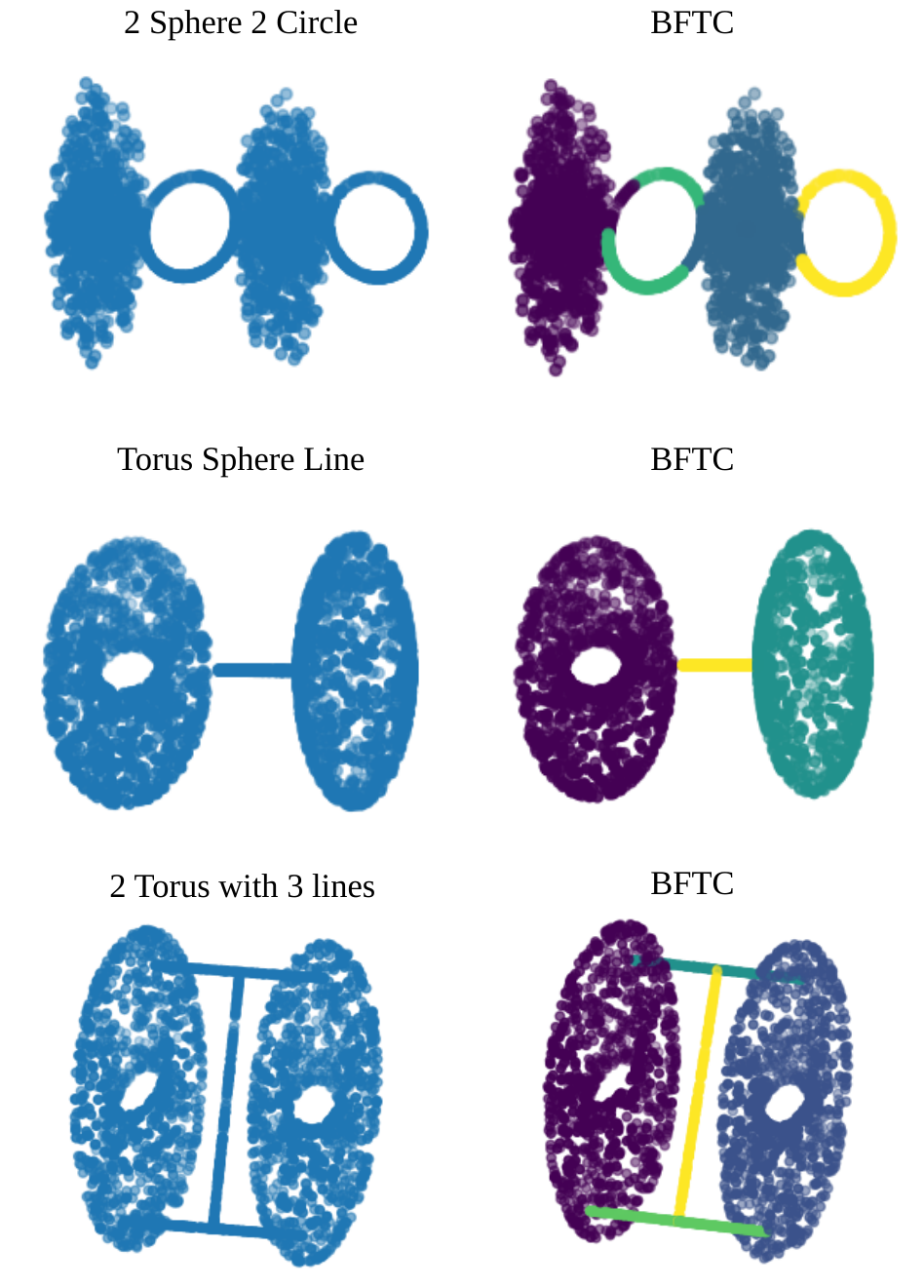}
    \caption{Performance of BFTC for $3$ Synthetic $3$D datasets.}
    \label{fig:synthetic_3d}
\end{figure}

\subsection{Experimental Setup and Baselines}
The performance of BFTC is measured using two well-recognized evaluation metrics: Adjusted Rand Index (ARI) \cite{ari} and Normalized Mutual Information (NMI) \cite{nmi}. The notable common benchmarks: $k$-Means \cite{kmeans}, Spectral \cite{spectral}, DBSCAN \cite{dbscan}, Mean Shift \cite{meanshift}, Optics \cite{optics}, Agglomerative \cite{agglomerative}, and Topological clustering benchmarks like ToMATo \cite{tomato}, TPCC \cite{tpcc}, CBN \cite{islambekov2019unsupervised}, TKM \cite{topokmeans} clustering algorithms are considered for the comparison with two variants of BFTC; $k$-NN\_BFTC and $\varepsilon\_$BFTC. For the faster computation of Betti numbers from VR-complex, we utilized the code base from the Ripser \cite{ripser}.

\subsection{Experiment on Synthetic Datasets}
We apply both variants of BFTC on three $2$-D and three $3$-D synthetic datasets. We also experiment with our method on the Linked Tori dataset, even with the presence of noise. The performance of our proposed method is presented in Table \ref{tab:synth_1}, comparing with several baselines. The visualization of well-formed clusters for $2$D and $3$D datasets are elucidated in Figures \ref{fig:synthetic_2d} and \ref{fig:synthetic_3d}. The data points contain complex geometrical structures, significantly complicating the formation of clusters by applying existing baselines. Both variants of BFTC run contrary by identifying the structural patterns that lead to the formation of clusters. Both visualization and numerical results underscore the benefits of exploiting topological information of the point sets, leading to well-formed clusters. 

\subsection{Experiments on Real Datasets}
BFTC is also applied to $7$ real-world datasets: Zoo, Ecoli, Wisconsin B.C., MNIST, Phishing URL, Glass, ORHD, and the performance is presented in Table \ref{tab:real}. The datasets contain a much higher number of data points and feature dimensions compared to the synthetic ones. Beyond the superior performance on synthetic datasets, we also achieved commendable performance on real-world datasets. The utility of the VR-complex and the Betti sequences identifies intricate structural similarities of the higher-dimensional points. Therefore, the numerical quantities reflect the effectiveness of exploring topological properties for clustering. 

\begin{figure}[h]
    \centering
    \includegraphics[width=0.6\textwidth]{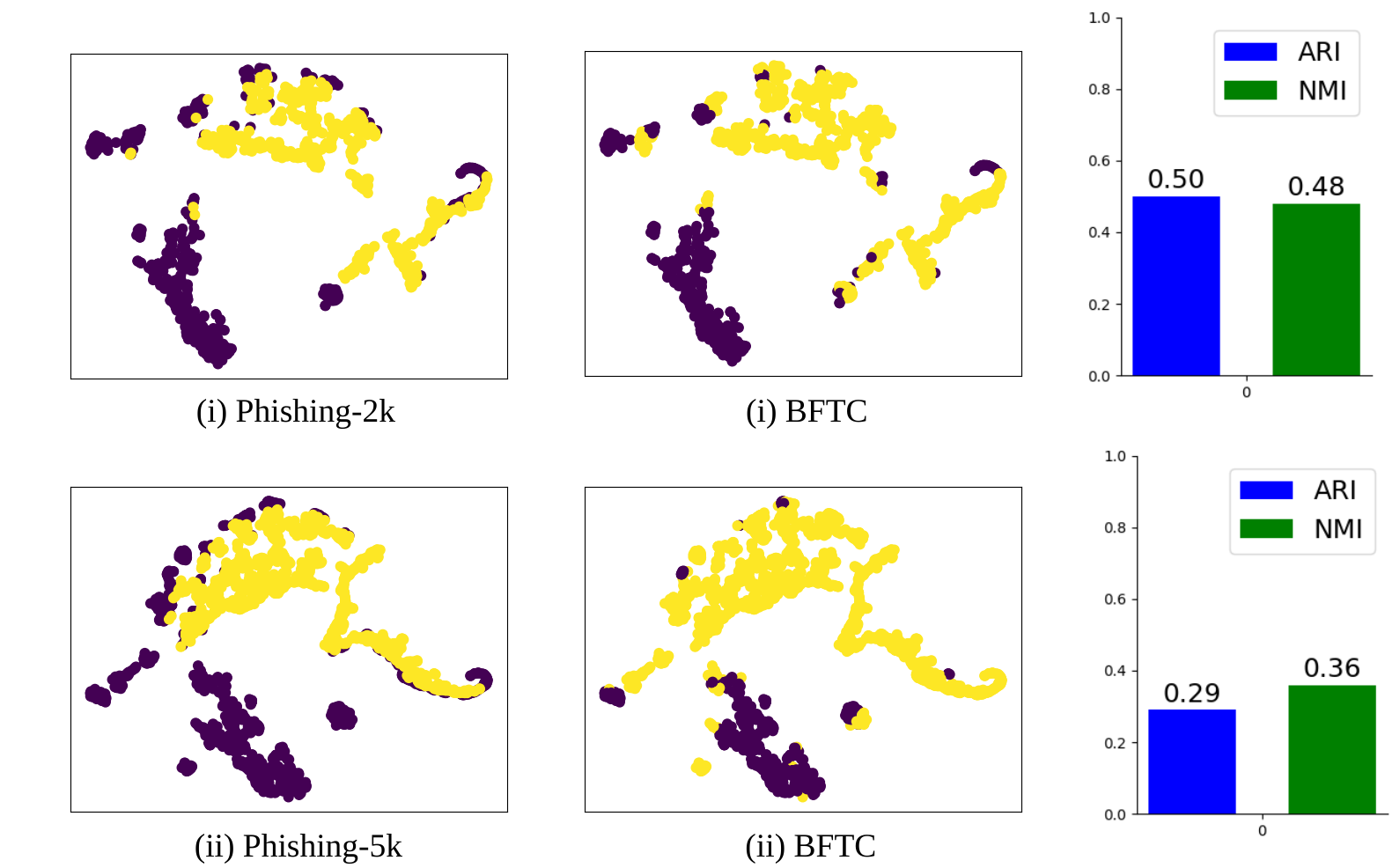}
    \caption{$k$-NN\_BFTC is applied to Phishing-2k and Phishing-5k datasets, and t-SNE visualizations of clusters are presented. The corresponding ARI and NMI values are also mentioned. }
    \label{fig:phising_two_settings}
\end{figure}

\subsection{More Performance Analyses}
We conduct experiments on the Phishing URL dataset in two different settings with $2k$ and $5k$ samples, respectively. Refer to Figure \ref{fig:phising_two_settings} to visualize the performance of $k$-NN\_BFTC on Phishing-2k and Phishing-5k. The corresponding values of ARI and NMI are also provided for each setting. The impressive performance of $k$-NN\_BFTC highlights the effectiveness of exploiting topological aspects of the data points. Beyond the limited number of points in the synthetic datasets, BFTC can further tackle larger numbers of data points with the formation of well-formed clusters. 

\subsection{Finding Thresholds}
The performance of both variants of BFTC is heavily reliant on choosing appropriate thresholds to identify topologically similar structures. For each $x_i \in X$, we defined a set of thresholds $\alpha_0, \alpha_1, \alpha_2, \cdots, \alpha_M$ acting as the lower bounds for the $(M+1)$ cosine similarity scores $(d_{ij})^0, (d_{ij})^1, (d_{ij})^{2}, \cdots, (d_{ij})^M$, respectively. These lower bounds are taken as the lower whisker values of the box plots for the corresponding cosine similarities, i.e. the first quartile (Q1) minus $1.5$ times the interquartile range (IQR), denoted as $\alpha_m = \{Q1 - 1.5 (Q3 - Q1)\}$, where $Q1$ is the $25^{th}$ percentile and $Q3$ is the $75^{th}$ percentile $\forall m \in [0, M]$ for different dimensional Betti numbers. Refer to Figure \ref{fig:boxplot_vs_betti} for the box plots for $\beta_0$ and $\beta_1$ of the Linked Tori dataset.

\begin{figure*}[!ht]
    \centering
    \includegraphics[height= 0.22\textheight, width= \linewidth]{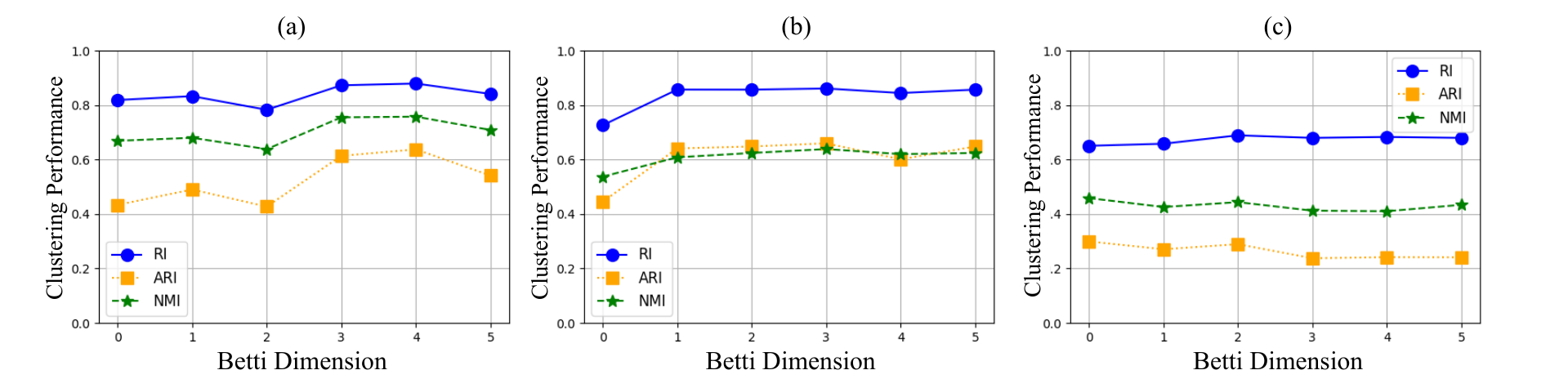}
    \caption{ The performance metrics RI, ARI, and NMI for (a) Zoo, (b) Ecoli, and (c) Glass datasets, respectively, with varying Betti numbers are presented. The Betti numbers are varied from dimension $0$ to $5$, and performance becomes optimal for certain Betti numbers that differ for every dataset.}
    \label{fig:betti_vs_perf}
\end{figure*}

\subsection{Performance of BFTC with Variation of Betti Numbers} 
We conduct an experiment to observe the effects of clustering by varying the dimension of Betti numbers in $k$-NN\_BFTC. Three real-world datasets, Zoo, Ecoli, and Glass are considered equipped with high-dimensional features. We have taken the Betti numbers from $0$ up to $5$-dimension and recorded the performance metrics for the respective datasets illustrated in Figure \ref{fig:betti_vs_perf}. The performance is assessed using three metrics: RI, ARI, and NMI. The metric values attain an optimal value for the specific Betti numbers, signifying the importance of selecting suitable Betti dimensions. Furthermore, the performance trends of the three metrics are almost similar, underscoring the consistency of the proposed algorithm.

\begin{figure}[h]
    \centering
    \includegraphics[width= 0.6\linewidth]{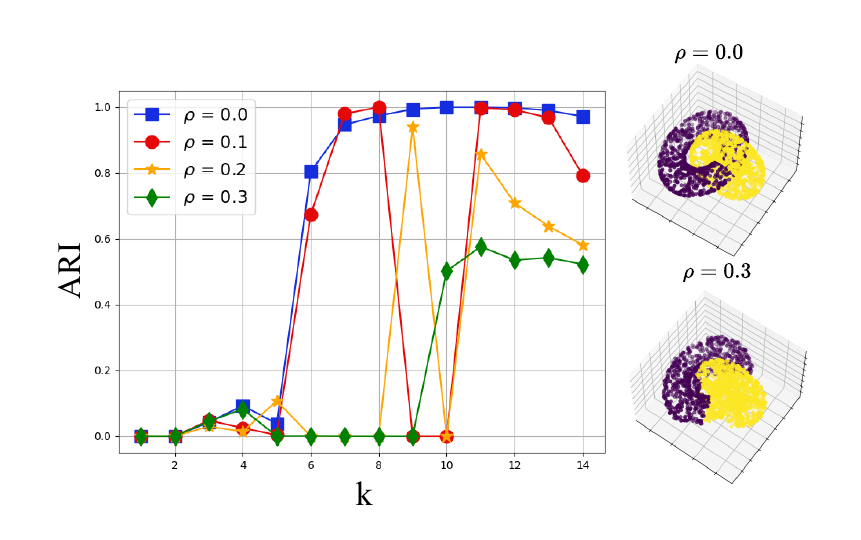}
    \caption{We have added i.i.d. Gaussian noise with varying standard deviation specified by the parameter $\rho$ on all three coordinates of every point. Left: Plot of ARI vs different values of $k$ in $k$-NN for different noise values $\rho = 0.10, 0.20, 0.30$ for the Linked Tori dataset. Right: Figure of clustering obtained by $k$-NN\_BFTC (ours) for the Linked Tori dataset with noise ($\rho = 0.0, \rho = 0.3$), respectively.}
    \label{fig:performance_vs_noise}
\end{figure}

\subsection{Robustness Analyses}
We have added Gaussian noise to the features of the Linked Tori dataset to study the resilience against noise perturbation. Three levels of Gaussian noise ($0.10$, $0.20$, and $0.30$) are added to compare the performance of $k$-NN\_BFTC with varying values of $k$. ARI is considered as the performance metric, and the corresponding plots are demonstrated in Figure \ref{fig:performance_vs_noise}. We have also mentioned the numerical results in Table \ref{tab:synth_1}. The results suggest that the performance of both variants of BFTC exhibited stability with the increasing noise level. Both variants of BFTC showed notable performance compared to the same without noise, even with the noise level of $0.30$. The performance trends indicate the robustness of BFTC against the noise. 

\begin{figure}
    \centering
    \includegraphics[width= 0.6\linewidth, height= 0.34\textheight]{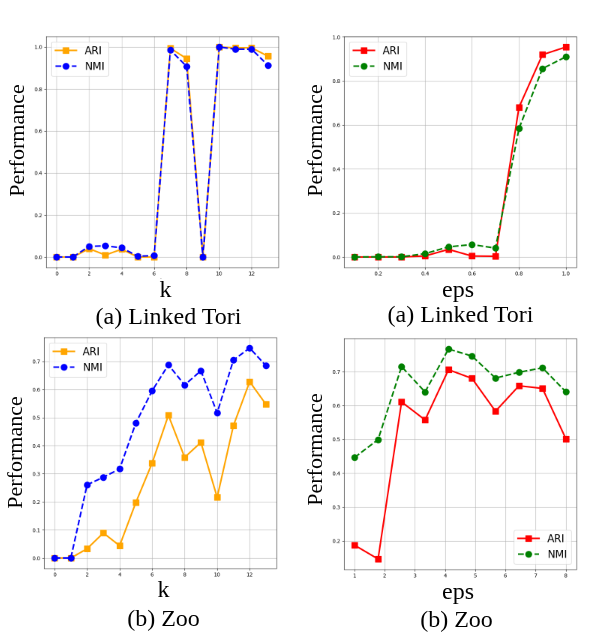}
    \caption{The performances of BFTC  for both the choice of neighborhoods, $k$-NN \& $\varepsilon$-neighborhood for (a) Linked Tori and (b) Zoo datasets.}
    \label{fig:nbd}
\end{figure}

\subsection{Ablation Study}
We conduct an ablation study by varying three key components, which are (1) the Neighborhood Construction, (2) the distance metric, and (3) the kernel function. 

\subsubsection{Neighborhood Construction.} To investigate the impact of neighborhood construction on topology-aware clustering, we conduct an ablation study on the synthetic \emph{Linked Tori} and a real dataset, Zoo, respectively. Due to the intertwined structure, local neighborhood selection plays a crucial role in preserving intrinsic topology.

We compare the following neighborhood construction strategies:

\begin{itemize}
    \item \textbf{$k$-NN\_BFTC:} Number of neighbors $k\in \{1,\dots,14\}$.
    \item \textbf{$\varepsilon\_$BFTC:} Radius $\varepsilon \in [0.1,1]$ for Linked Tori and $\varepsilon \in [1,8]$ for Zoo.
\end{itemize}

Figure~\ref{fig:nbd} illustrates clustering performance (ARI and NMI) as a function of
neighborhood parameters for both the Linked Tori and Zoo datasets.
The left column corresponds to the $k$-NN construction,
while the right column corresponds to the $\varepsilon$-neighborhood construction.

For the Linked Tori dataset, performance under the $k$-NN construction exhibits sharp transitions.
Very small values of $k$ result in poor clustering accuracy,
as the local neighborhoods are too sparse to capture the intrinsic toroidal structure. Once $k$ exceeds a critical threshold, ARI and NMI rapidly increase and approach near-perfect clustering.
However, further increasing $k$ introduces fluctuations,
indicating the emergence of shortcut connections between the linked manifolds. Under the $\varepsilon$-neighborhood construction,
performance remains low for small $\varepsilon$ due to disconnected local graphs. As $\varepsilon$ increases, performance improves sharply once the filtration scale becomes sufficient to capture the intrinsic loops of each torus. Beyond this regime, performance stabilizes at high values, reflecting the successful separation of the two linked components.

For the Zoo dataset, the dependence on $k$ is smoother. Clustering performance gradually improves as $k$ increases,
suggesting that richer neighborhood information enhances the discrimination of categorical feature patterns.
However, excessively small neighborhoods fail to provide sufficient
topological context.

The linked Tori dataset contains non-trivial first homology ($\beta_1$) corresponding to toroidal loops. Accurate recovery of these features requires neighborhoods that are large enough to capture intrinsic cycles but sufficiently local enough to avoid cross-manifold shortcuts. The $k$-NN graph enforces bounded degree and controlled sparsity, yielding smoother spectral behavior and more stable Betti sequences. In contrast, the $\varepsilon$-neighborhood graph enforces a fixed metric scale, which may not uniformly align with the intrinsic manifold curvature across the dataset. For the Zoo dataset, where topology primarily reflects
connectivity patterns in feature space,
moderate neighborhood scales enhance discrimination of structural patterns without inducing spurious cycles.
Thus, optimal performance occurs when the neighborhood scale
aligns with the intrinsic geometric and topological scale of the data.

This ablation demonstrates that neighborhood construction significantly influences both topological feature extraction and final clustering performance on geometrically complex datasets. 

\begin{figure}
    \centering
    \includegraphics[width= 0.6\linewidth]{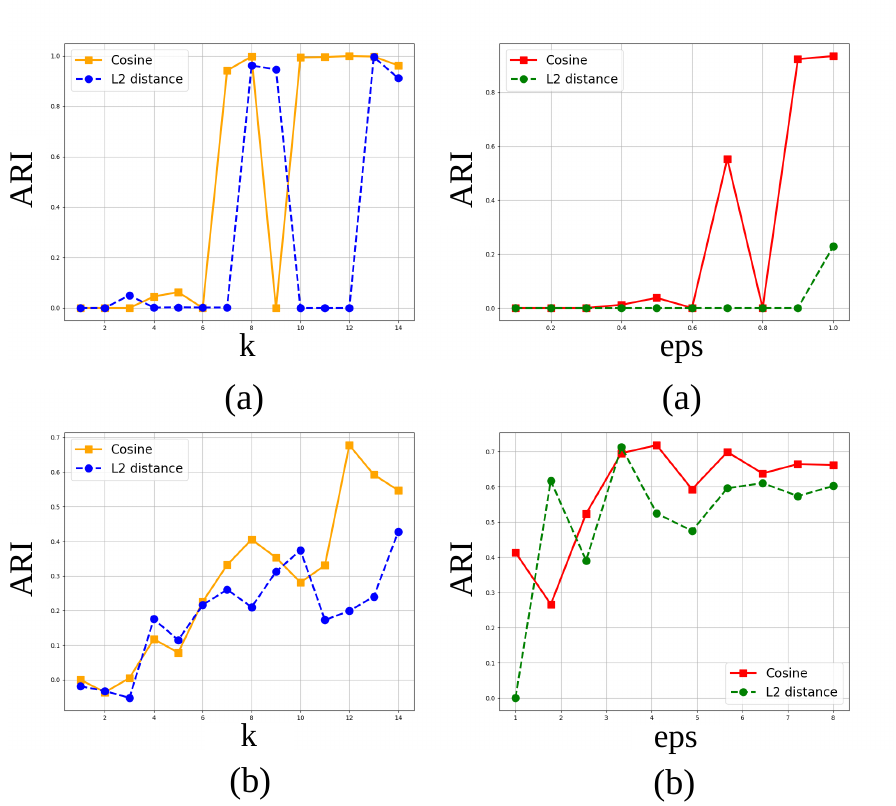}
    \caption{The performances of two variants of BFTC for both distance metrics, Cosine Similarity \& L2-norm, are presented for (a) Linked tori and (b) Zoo datasets.}
    \label{fig:distance}
\end{figure}

\begin{figure}
    \centering
    \includegraphics[width= 0.6\linewidth]{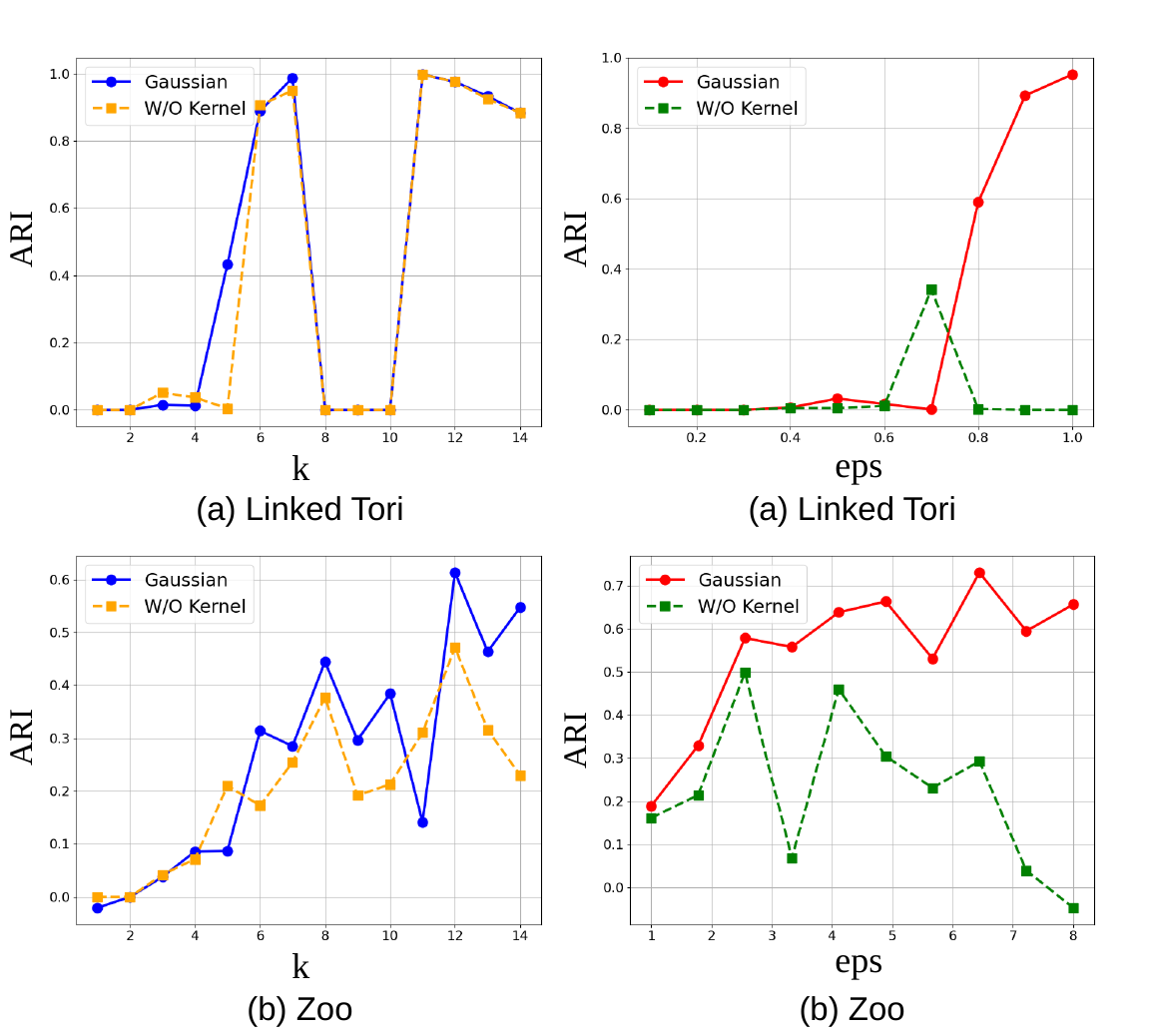}
    \caption{The performances of two variants of BFTC are presented for both cases of using a Gaussian kernel and without using any kernel function.}
    \label{fig:kernel}
\end{figure}

\subsubsection{Choice of Distance Metrics.} To evaluate the impact of similarity measures on topology-aware clustering, we conduct experiments using two different distance metrics:
the Euclidean ($\ell_2$) distance and the cosine similarity as defined in Eq.~\ref{eq:cosine}. Both metrics are examined under the $k$-NN and $\varepsilon$-neighbor constructions. The parameter $k$ is varied from $1$ to $14$, while $\varepsilon$ is varied across the admissible range for each dataset.
The resulting ARI scores are shown in Figure~\ref{fig:distance}. For the Linked Tori dataset, cosine similarity consistently achieves
near-perfect clustering performance once the neighborhood parameter
exceeds a moderate threshold, under both $k$ and $\varepsilon$ constructions. In contrast, the  $\ell_2$ distance exhibits unstable behavior, with significantly lower ARI values across most parameter ranges. This indicates that cosine similarity better preserves the intrinsic topological relationships of the intertwined manifolds. For the Zoo dataset, the performance gap remains evident.
Under the $k$-NN construction, cosine similarity produces steadily improving ARI scores as $k$ increases, while $\ell_2$ distance shows weaker and more irregular performance. Similarly, under the $\varepsilon$-neighborhood construction, cosine similarity yields consistently higher ARI values across a wide range of $\varepsilon$, whereas $\ell_2$ distance remains comparatively less discriminative.

Overall, the results demonstrate that cosine similarity more effectively captures structural similarity between Betti sequences than Euclidean distance. Since Betti sequences encode multiscale topological patterns rather than absolute magnitudes, angular similarity provides a more robust comparison.
Consequently, employing cosine similarity significantly enhances the clustering performance of BFTC across both geometric (Linked Tori) and tabular (Zoo) datasets.

\subsubsection{Choice of Kernels.} We conduct two separate experiments with the two variants of BFTC, one by applying a Gaussian kernel in forming the adjacency matrix $\mathcal{A}'$ as described in Eq. \ref{eq:kernel} and another without employing any kernel. The performance is presented in Figure \ref{fig:kernel}. The trends suggest that involving the Gaussian kernel improves the performance of both the variants of BFTC compared to the same when no kernel is applied. The kernel function assigns weights, making it easier to distinguish identical points and leading to better cluster formation. 

\section{Conclusion \& Future works}
In this work, we proposed a novel topological clustering algorithm, BFTC, that integrates multiscale topological information into the clustering process through Vietoris–Rips filtrations and Betti sequences. By encoding the evolution of topological features across filtration scales, both variants of BFTC effectively capture both local geometric structure and higher-dimensional topology, enabling robust clustering of datasets with complex, nonconvex, and intertwined structures. The proposed topology-aware graph refinement further enhances cluster separability while reducing redundant neighborhood connections, leading to improved computational efficiency. Extensive experiments on the synthetic and real-world datasets demonstrate that both variants of BFTC consistently outperform existing methods, including topology-based baselines, in terms of clustering accuracy and robustness. Designing clustering algorithms by investigating topological aspects for large-scale datasets with heavily complex overlapping structures and exploring adaptive filtration strategies can be a potential avenue for future research direction.

\bibliographystyle{IEEETran}
\bibliography{reference}

\begin{thebibliography}{10}
\providecommand{\url}[1]{#1}
\csname url@samestyle\endcsname
\providecommand{\newblock}{\relax}
\providecommand{\bibinfo}[2]{#2}
\providecommand{\BIBentrySTDinterwordspacing}{\spaceskip=0pt\relax}
\providecommand{\BIBentryALTinterwordstretchfactor}{4}
\providecommand{\BIBentryALTinterwordspacing}{\spaceskip=\fontdimen2\font plus
\BIBentryALTinterwordstretchfactor\fontdimen3\font minus \fontdimen4\font\relax}
\providecommand{\BIBforeignlanguage}[2]{{%
\expandafter\ifx\csname l@#1\endcsname\relax
\typeout{** WARNING: IEEEtran.bst: No hyphenation pattern has been}%
\typeout{** loaded for the language `#1'. Using the pattern for}%
\typeout{** the default language instead.}%
\else
\language=\csname l@#1\endcsname
\fi
#2}}
\providecommand{\BIBdecl}{\relax}
\BIBdecl

\bibitem{zomorodian2012tda}
A.~Zomorodian, ``Topological data analysis,'' \emph{Advances in applied and computational topology}, vol.~70, pp. 1--39, 2012.

\bibitem{tda}
C.~Epstein, G.~Carlsson, and H.~Edelsbrunner, ``Topological data analysis,'' \emph{Inverse Problems}, vol.~27, no.~12, p. 120201, 2011.

\bibitem{bois2024persistence}
A.~Bois, B.~Tervil, and L.~Oudre, ``Persistence-based clustering with outlier-removing filtration,'' \emph{Frontiers in Applied Mathematics and Statistics}, vol.~10, p. 1260828, 2024.

\bibitem{kmeans}
T.~Kanungo, D.~M. Mount, N.~S. Netanyahu, C.~D. Piatko, R.~Silverman, and A.~Y. Wu, ``An efficient k-means clustering algorithm: Analysis and implementation,'' \emph{IEEE transactions on pattern analysis and machine intelligence}, vol.~24, no.~7, pp. 881--892, 2002.

\bibitem{spectral}
U.~Von~Luxburg, ``A tutorial on spectral clustering,'' \emph{Statistics and computing}, vol.~17, pp. 395--416, 2007.

\bibitem{meanshift}
Y.~Cheng, ``Mean shift, mode seeking, and clustering,'' \emph{IEEE transactions on pattern analysis and machine intelligence}, vol.~17, no.~8, pp. 790--799, 1995.

\bibitem{dbscan}
E.~Schubert, J.~Sander, M.~Ester, H.~P. Kriegel, and X.~Xu, ``Dbscan revisited, revisited: why and how you should (still) use dbscan,'' \emph{ACM Transactions on Database Systems (TODS)}, vol.~42, no.~3, pp. 1--21, 2017.

\bibitem{optics}
M.~Ankerst, M.~M. Breunig, H.-P. Kriegel, and J.~Sander, ``Optics: Ordering points to identify the clustering structure,'' \emph{ACM Sigmod record}, vol.~28, no.~2, pp. 49--60, 1999.

\bibitem{agglomerative}
D.~M{\"u}llner, ``Modern hierarchical, agglomerative clustering algorithms,'' \emph{arXiv preprint arXiv:1109.2378}, 2011.

\bibitem{tomato}
F.~Chazal, L.~J. Guibas, S.~Y. Oudot, and P.~Skraba, ``Persistence-based clustering in riemannian manifolds,'' \emph{Journal of the ACM (JACM)}, vol.~60, no.~6, pp. 1--38, 2013.

\bibitem{tpcc}
V.~P. Grande and M.~T. Schaub, ``Topological point cloud clustering,'' \emph{arXiv preprint arXiv:2303.16716}, 2023.

\bibitem{topokmeans}
M.~Dixon, Y.~Chen, and Y.~R. Gel, ``Topological k-means clustering in reproducing kernel hilbert spaces,'' \emph{Electronic Journal of Statistics}, vol.~19, no.~1, pp. 204--239, 2025.

\bibitem{knn}
L.~E. Peterson, ``K-nearest neighbor,'' \emph{Scholarpedia}, vol.~4, no.~2, p. 1883, 2009.

\bibitem{wasserman2018topological}
L.~Wasserman, ``Topological data analysis,'' \emph{Annual Review of Statistics and Its Application}, vol.~5, no.~1, pp. 501--532, 2018.

\bibitem{zomorodian2004computing}
A.~Zomorodian and G.~Carlsson, ``Computing persistent homology,'' in \emph{Proceedings of the twentieth annual symposium on Computational geometry}, 2004, pp. 347--356.

\bibitem{edelsbrunner2002topological}
Edelsbrunner, Letscher, and Zomorodian, ``Topological persistence and simplification,'' \emph{Discrete \& computational geometry}, vol.~28, pp. 511--533, 2002.

\bibitem{panagopoulos2022topological}
D.~Panagopoulos, ``Topological data analysis and clustering,'' \emph{arXiv preprint arXiv:2201.09054}, 2022.

\bibitem{islambekov2019unsupervised}
U.~Islambekov and Y.~R. Gel, ``Unsupervised space--time clustering using persistent homology,'' \emph{Environmetrics}, vol.~30, no.~4, p. e2539, 2019.

\bibitem{majumdar2020clustering}
S.~Majumdar and A.~K. Laha, ``Clustering and classification of time series using topological data analysis with applications to finance,'' \emph{Expert Systems with Applications}, vol. 162, p. 113868, 2020.

\bibitem{yuvarajtopological}
M.~Yuvaraj, A.~K. Dey, V.~Lyubchich, Y.~R. Gel, and H.~V. Poor, ``Topological clustering of multilayer networks,'' \emph{Proceedings of the National Academy of Sciences}, vol. 118, no.~21, p. e2019994118, 2021.

\bibitem{alpha}
N.~Akkiraju, H.~Edelsbrunner, M.~Facello, P.~Fu, E.~Mucke, and C.~Varela, ``Alpha shapes: definition and software,'' in \emph{Proceedings of the 1st international computational geometry software workshop}, vol.~63, no.~66, 1995.

\bibitem{cech}
S.~Dantchev and I.~Ivrissimtzis, ``Efficient construction of the {\v{c}}ech complex,'' \emph{Computers \& Graphics}, vol.~36, no.~6, pp. 708--713, 2012.

\bibitem{vrcomplex}
A.~Zomorodian, ``Fast construction of the vietoris-rips complex,'' \emph{Computers \& Graphics}, vol.~34, no.~3, pp. 263--271, 2010.

\bibitem{ripser}
U.~Bauer, ``Ripser: efficient computation of vietoris--rips persistence barcodes,'' \emph{Journal of Applied and Computational Topology}, vol.~5, no.~3, pp. 391--423, 2021.

\bibitem{lanczos}
J.~K. Cullum and R.~A. Willoughby, \emph{Lanczos algorithms for large symmetric eigenvalue computations: Vol. I: Theory}.\hskip 1em plus 0.5em minus 0.4em\relax SIAM, 2002.

\bibitem{dua2017uci}
D.~Dua, C.~Graff \emph{et~al.}, ``Uci machine learning repository, 2017,'' \emph{URL http://archive. ics. uci. edu/ml}, vol.~7, no.~1, p.~62, 2017.

\bibitem{ari}
J.~M. Santos and M.~Embrechts, ``On the use of the adjusted rand index as a metric for evaluating supervised classification,'' in \emph{International conference on artificial neural networks}.\hskip 1em plus 0.5em minus 0.4em\relax Springer, 2009, pp. 175--184.

\bibitem{nmi}
P.~A. Est{\'e}vez, M.~Tesmer, C.~A. Perez, and J.~M. Zurada, ``Normalized mutual information feature selection,'' \emph{IEEE Transactions on neural networks}, vol.~20, no.~2, pp. 189--201, 2009.

\end{thebibliography}

\end{document}